\documentclass[letterpaper,journal]{IEEEtran}
\usepackage{amsmath,amsfonts}
\usepackage{algorithmic}
\usepackage{algorithm}
\usepackage{array}
\usepackage[caption=false,font=normalsize,labelfont=sf,textfont=sf]{subfig}
\usepackage{textcomp}
\usepackage{stfloats}
\usepackage{url}
\usepackage{verbatim}
\usepackage{graphicx}
\usepackage{cite}
\usepackage{hyperref} 
\usepackage{booktabs}
\usepackage[table]{xcolor}
\usepackage[textsize=tiny]{todonotes}
\newcommand{\eg}{e.g.,~}

\usepackage{enumitem}
\setlist[itemize]{nosep}

\usepackage{multirow}
\usepackage{colortbl}
\usepackage{xcolor}
\usepackage{pifont}
\newcommand{\cmark}{\textcolor{green!60!black}{\ding{51}}} 
\newcommand{\xmark}{\textcolor{red}{\ding{55}}}            

\usepackage{algorithm, algorithmic}

\definecolor{lightcyan}{RGB}{180,240,240}

\hypersetup{
    colorlinks,
    linkcolor={blue},
    urlcolor={magenta}
}

\begin{document}


\title{EduVQA: Towards Concept-Aware Assessment of Educational AI-Generated Videos}

\author{
Baoliang Chen*,~\IEEEmembership{Member,~IEEE},
Xinlong Bu*,
Hanwei Zhu,
Lingyu Zhu,
and Jieyu Zhan,~\IEEEmembership{Member,~IEEE}

\IEEEcompsocitemizethanks{
\IEEEcompsocthanksitem *These authors contributed equally.
\IEEEcompsocthanksitem
Baoliang Chen is with the College of Computing and Data Science,
Nanyang Technological University.
(E-mail: blchen6-c@my.cityu.edu.hk.)
\IEEEcompsocthanksitem
Xinlong Bu and Jieyu Zhan are with the Department of Computer Science,
South China Normal University, China.
(E-mails: xlbu@m.scnu.edu.cn; 	zhanjieyu@scnu.edu.cn.)
\IEEEcompsocthanksitem
Hanwei Zhu and W. Lin are with the College of Computing and Data Science,
Nanyang Technological University.
(E-mails: hanwei.zhu@ntu.edu.sg; wslin@ntu.edu.sg.)
\IEEEcompsocthanksitem
Lingyu Zhu is with the School of Computer Science,
City University of Hong Kong.
(E-mail: lingyzhu-c@my.cityu.edu.hk.)
\IEEEcompsocthanksitem
Corresponding author: Jieyu Zhan.
}
}

\maketitle

\begin{abstract}
Existing AI-generated video quality assessment (AIGVQA) methods mainly focus on global perceptual realism and coarse text-video alignment, while overlooking a critical requirement in educational scenarios: \emph{concept correctness}. In early mathematics education, subtle errors in numerical quantities, geometric relations, or spatial configurations may fundamentally alter the conveyed knowledge despite visually plausible generation. To address this problem, we introduce \textbf{EduAVQABench}, the first benchmark for concept-aware educational AIGV assessment, containing 1,130 videos generated by ten state-of-the-art T2V models together with over 310,650 fine-grained human annotations spanning perceptual quality and semantic alignment.
Built upon this benchmark, we further propose \textbf{EduVQA}, a concept-aware AIGVQA framework equipped with a Structured 2D Mixture-of-Experts (S2D-MoE) architecture. By jointly modeling fine-grained concept assessment and overall quality prediction through shared experts and adaptive two-dimensional routing, EduVQA effectively captures subtle concept-level inconsistencies overlooked by conventional global scoring methods. Extensive experiments demonstrate that EduVQA consistently outperforms existing AIGVQA approaches across both perceptual and semantic evaluation tasks while exhibiting strong generalization capability on unseen benchmarks. 
Code and dataset will be publicly available at: \url{https://github.com/EduVQA/EduVQA}.
\end{abstract}


\section{Introduction}
Recent advances in text-to-video (T2V) generation models~\cite{Li2018VideoGenFromText,Hong2022CogVideo,OpenAI2024Sora,kling} have significantly improved the realism and controllability of AI-generated videos (AIGVs). Beyond entertainment and media creation, such progress also provides new opportunities for educational content generation, particularly for scenarios that benefit from intuitive visual demonstrations. Among them, early mathematics education serves as a representative application, since many foundational concepts (\eg counting, geometric relations, spatial transformations, and measurement) rely heavily on precise visual grounding for effective understanding.

\begin{figure}[t]
\centering
\includegraphics[width=1.0\linewidth]{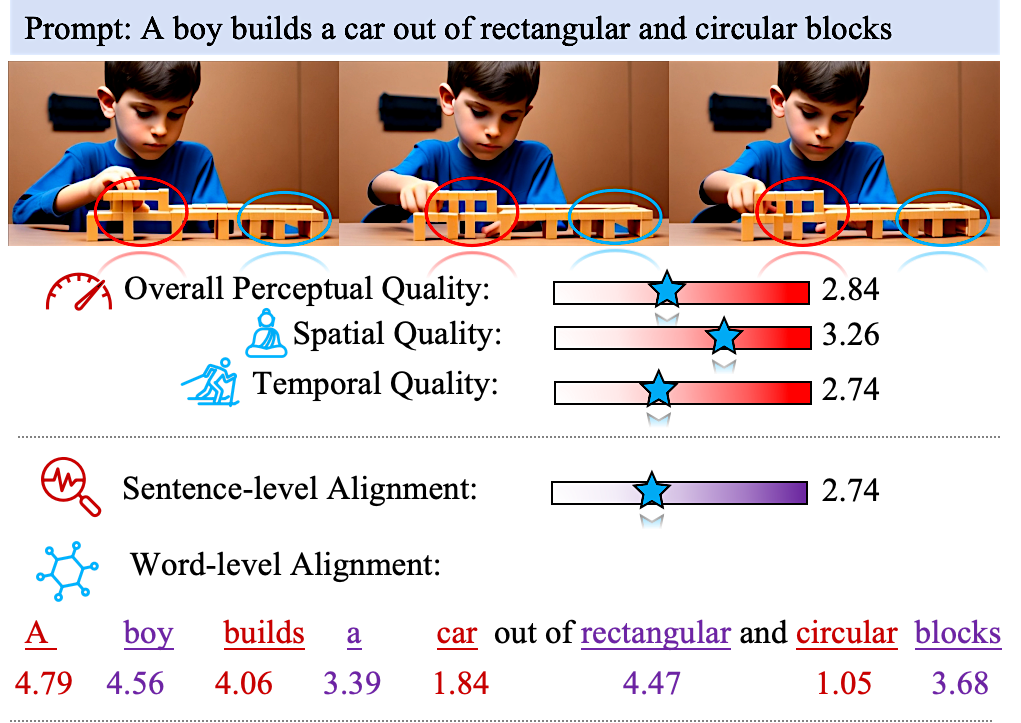}  
\vspace{-5mm}
\caption{Annotation structure of our constructed EduAVQABench dataset. Each educational video is annotated with spatial and temporal fidelity and word-level semantic consistency, enabling fine-grained assessment of perceptual quality and prompt alignment. The red and blue elliptical regions indicate temporal inconsistencies that negatively impact temporal quality.}
\vspace{-5mm}
\label{fig:annotation}
\end{figure}

Despite this potential, the evaluation of educational AIGVs remains largely unexplored. Existing AIGV quality assessment (AIGVQA) methods mainly focus on global perceptual realism or overall text-video alignment~\cite{kou2024subjective,lu2024aigc}, typically predicting a single quality score or several coarse semantic metrics. Such paradigms are suitable for entertainment-oriented videos, where overall visual plausibility is the primary objective. However, educational AIGVs require a fundamentally different evaluation criterion: \textit{\textbf{whether each underlying concepts are correctly represented}}.

For example, a generated video for the prompt ``\textit{three blue square blocks}'' may appear visually realistic while containing incorrect object counts, shapes, or colors. Although these local inconsistencies may only marginally affect overall perceptual realism, they directly alter the conveyed educational content. Similar issues frequently arise in early mathematics scenarios involving numerical quantities, geometric attributes, spatial relations, or measurement concepts. These observations reveal a key limitation of existing AIGVQA methods: global quality scores often average out fine-grained concept errors, making them insufficient for educationally meaningful assessment.

To address this problem, we first introduce \textbf{EduAVQABench}, a benchmark dedicated to educational AIGV assessment in early mathematics scenarios. EduAVQABench contains 1,130 videos generated by ten state-of-the-art T2V models using 113 expert-curated prompts covering four foundational domains: \textit{Numbers}, \textit{Geometry}, \textit{Measurement}, and \textit{Probability}. More importantly, unlike existing AIGVQA datasets relying mainly on coarse global annotations, our benchmark provides fine-grained annotations from both local and global perspectives, including spatial quality, temporal quality, overall perceptual quality, word-level alignment, and sentence-level alignment. Such annotations explicitly characterize whether individual educational concepts are correctly grounded in generated videos, enabling concept-aware quality assessment. An overview of our annotation framework is illustrated in Fig.~\ref{fig:annotation}.

Built upon this benchmark, we further propose \textbf{EduVQA}, a concept-aware AIGVQA framework based on a \textbf{Structured 2D Mixture-of-Experts (S2D-MoE)} architecture. Instead of only predicting holistic quality scores, EduVQA jointly models fine-grained sub-dimensional assessment and overall quality prediction. Specifically, the framework adopts a dual-path architecture, where the perceptual path evaluates spatial-temporal quality while the alignment path models word-level and sentence-level semantic consistency. 

A key challenge is that concept-aware evaluation naturally involves hierarchical quality dependencies: overall educational quality is intrinsically influenced by multiple fine-grained perceptual and semantic factors. Conventional MoE architectures usually treat different prediction targets independently, lacking mechanisms to explicitly model the relationship between local concept evaluation and global quality assessment. To address this issue, the proposed S2D-MoE introduces shared expert representations together with adaptive two-dimensional routing, enabling collaborative learning between overall and sub-dimensional predictions. Such a design allows the model to better capture subtle concept-level errors while simultaneously improving interpretability and generalization capability.

Extensive experiments demonstrate that EduVQA consistently outperforms existing AIGVQA baselines on both perceptual quality assessment and semantic alignment evaluation. Moreover, qualitative analyses show that our method can effectively identify fine-grained concept inconsistencies that are often overlooked by conventional global scoring methods.

Our contributions are summarized as follows:

\begin{itemize}

\item We introduce EduAVQABench, the first benchmark for educational AIGV assessment in early mathematics scenarios, containing 1,130 videos generated by ten state-of-the-art T2V models together with over 310,650 fine-grained human annotations.

\item We propose a concept-aware annotation protocol that jointly evaluates perceptual quality and fine-grained semantic alignment from both local and global perspectives, enabling interpretable educational AIGV assessment beyond conventional holistic quality scoring.

\item We propose EduVQA, a concept-aware AIGVQA framework equipped with a dual-path Structured 2D Mixture-of-Experts (S2D-MoE) architecture, which explicitly models the hierarchical relationship between fine-grained concept assessment and overall quality prediction.

\item Extensive experiments demonstrate that EduVQA achieves state-of-the-art performance on educational AIGV assessment and exhibits strong generalization capability across multiple unseen AIGVQA benchmarks.

\end{itemize}

\section{Related Work}

\subsection{Video Generation Models}

T2V generation has rapidly evolved through three architectural paradigms. VAE/GAN‑based models~\cite{Li2018VideoGenFromText,Deng2019IRCGAN} extend image VAEs and GANs by adding temporal modules (\eg 3D convolutions or recurrent layers) to synthesize short clips. While capable of fast sampling, these approaches often suffer from mode collapse and limited motion coherence. Autoregressive models treat frame generation as a sequence prediction task, leveraging transformer architectures to predict pixels or latent tokens one step at a time~\cite{Wu2022NUWA,Hong2022CogVideo}. This yields improved fidelity but tends to accumulate errors and exhibit temporal drift over long horizons. Diffusion models have recently become dominant, generating videos by iteratively denoising noise conditioned on spatial–temporal context \cite{Ho2022VideoDiffusion,Chen2023VideoCrafter1,Esser2023Gen1,OpenAI2024Sora,kling}. By refining all frames jointly at each step, diffusion methods avoid adversarial instability and deliver superior global consistency. 
Despite these advances, existing T2V models still struggle with content fidelity and text–video alignment, underscoring the need for dedicated AIGVQA supervision to guide quality-aware video generation.

\subsection{Quality Metrics for AIGC Videos}
Assessing AIGVs requires metrics that capture both perceptual quality (spatial and temporal fidelity) and prompt alignment (semantic correctness). Early evaluations borrowed image‑centric metrics like Inception Score (IS) \cite{salimans2016improved} 
alongside classical IQA measures (\eg NIQE~\cite{mittal2012making}) and learned predictors (\eg StairIQA~\cite{sun2023blind}, TOPIQ~\cite{chen2024topiq}, Grounding-IQA~\cite{chen2024grounding}, PRL-Net~\cite{chen2026toward}, LSGS~\cite{zhou2023blind}). However, these frame‑level approaches ignore temporal artifacts and loosely correlate with human judgments. Video‑level extensions such as Frechet Video Distance (FVD)~\cite{Unterthiner2019FVD} and UGC‑focused VQA models (\eg VSFA~\cite{li2019quality}, SAMA~\cite{liu2024scaling}, CM2PL~\cite{chen2025exploring}, MGQA~\cite{yan2024video}, NR-VQA~\cite{shen2023blind}, KSVQE~\cite{lu2024kvq}) incorporate motion consistency but remain trained on natural videos, limiting sensitivity to AIGC‑specific distortions. For semantic alignment, most methods employ CLIP‑based similarity by averaging frame‑wise text-image scores~\cite{Radford2021CLIP}, which overlooks concept order and dynamic content. Benchmarks like EvalCrafter~\cite{liu2024evalcrafter}, AIGC-VQA~\cite{lu2024aigc}, and VBench~\cite{huang2024vbench} propose multiple objective metrics, yet still lack fine‑grained evaluation for domains such as educational video generation.

\begin{figure*}[t]
  \centering
  \includegraphics[width=\textwidth]{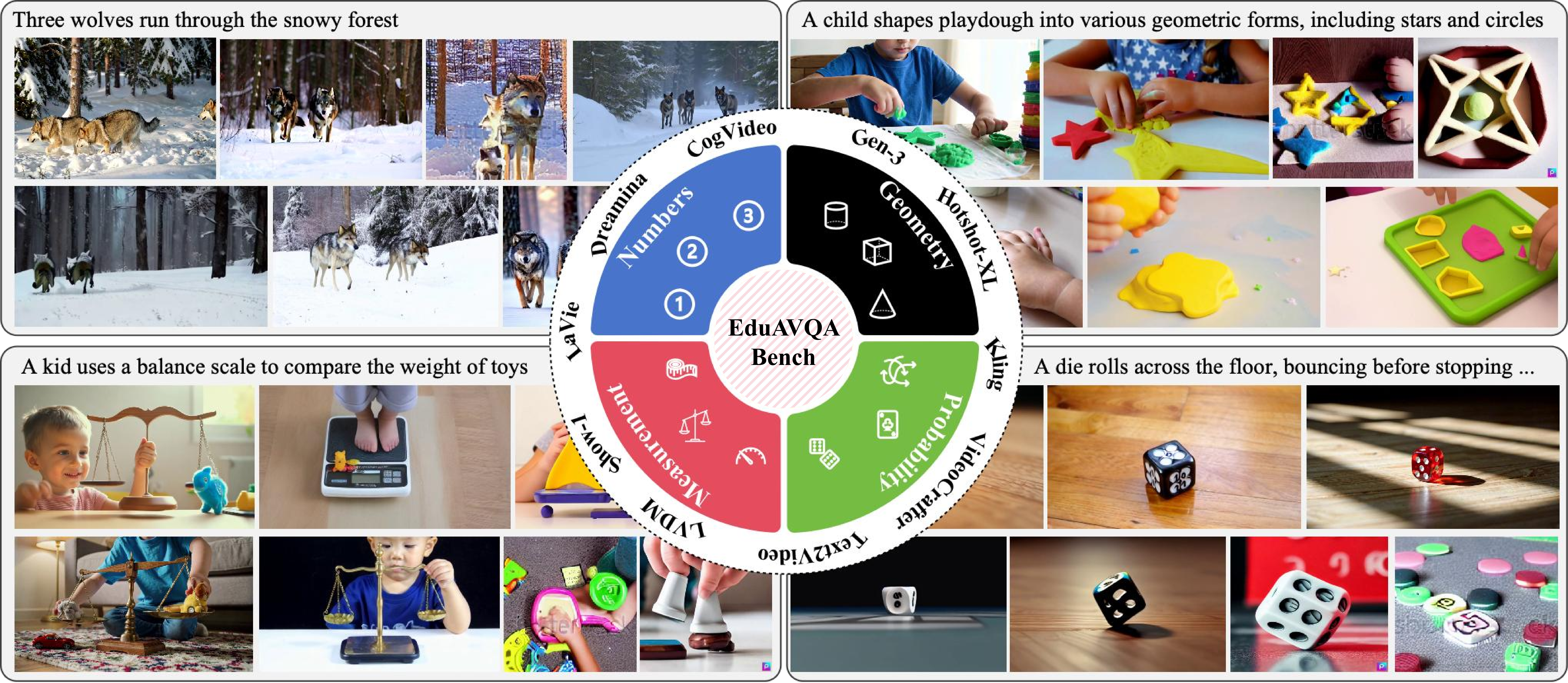}
  \vspace{-5mm}
 \caption{An overview of our dataset, divided into four categories: \textit{Numbers}, \textit{Geometry}, \textit{Measurement}, and \textit{Probability}.}
 \vspace{-5mm}
  \label{fig:dataset}
\end{figure*}

\section{EduAVQABench: Fine-Grained Video Quality Evaluation in Early Math Education}

\subsection{Data Collection}
Our {EduAVQABench} dataset is constructed to facilitate fine-grained evaluation of AIGVs in the context of {mathematics education}, with a particular emphasis on {young learners}. Although recent T2V models have achieved remarkable progress, they still struggle to render highly \textit{abstract, symbolic, or spatially precise} concepts. For instance, most models fail to accurately depict complex numerosity (\eg ``fifteen children dancing in a circle'') or detailed geometric transformations (\eg ``a triangle rotating 90 degrees clockwise around its vertex''). Such precise object counting and subtle motion representation remain beyond current T2V capabilities. Given these limitations, although our long-term goal is to support a broad range of mathematical content, the initial version of our dataset is intentionally focused on visually realizable concepts that current T2V models can reasonably depict. These concepts, in turn,  are particularly well-suited for young learners, who benefit most from the visual grounding of abstract ideas.

\noindent\textbf{Prompt Design.}
To guarantee pedagogical relevance, we derive prompt content from widely recognized educational standards: the \emph{TIMSS 2023 Assessment Frameworks~\cite{mullis2021timss}} and the \emph{Common Core State Standards for Mathematics} (CCSSI, United States~\cite{core2010common}). Based on these sources, we identify four domains where current T2V capabilities are most applicable: \textit{Numbers}, \textit{Geometry}, \textit{Measurement}, and \textit{Probability}. We leverage GPT-4o to generate candidate prompts by grounding curriculum-relevant concepts in concrete, visually plausible everyday scenarios, bridging the gap between abstract mathematical ideas and T2V generation capability. 
From an initial pool, we curate {113 high-quality prompts} via expert filtering, covering four core domains in early mathematics education: \textit{Numbers (43 prompts)}, \textit{Geometry (40 prompts)}, \textit{Measurement (20 prompts)}, and \textit{Probability (10 prompts)}, ensuring topical diversity. Each domain encompasses several pedagogically meaningful subtypes to balance conceptual breadth and visual renderability. Specifically, the Numbers domain spans key concepts such as \textit{counting, number comparison, basic number operations, and basic fractions}; Geometry includes \textit{shape recognition, shape composition, spatial reasoning, and reasoning about attributes}; Measurement focuses on \textit{measurable attributes and categorical data}; and Probability introduces \textit{foundational ideas of randomness and chance}, which are essential for early probabilistic thinking. Example videos across the four topics are visualized  in Fig.~\ref{fig:dataset}.

\noindent\textbf{T2V Model Selection.}
To ensure diversity and representativeness in the generated content, we employ {ten widely-used T2V models}, encompassing both open-source and commercial systems. The selected models include: \emph{CogVideo}~\cite{Hong2022CogVideo}, \emph{Gen-3}~\cite{runway2024gen3}, \emph{Hotshot-XL}~\cite{Mullan_Hotshot_XL_2023}, \emph{Dreamina}~\cite{dreamina2024}, \emph{Kling~\cite{kling}}, \emph{LaVie}~\cite{wang2023lavie}, \emph{LVDM}~\cite{he2022latent}, \emph{Show-1}~\cite{zhang2025show}, \emph{Text2Video-Zero}~\cite{khachatryan2023text2video}, and \emph{VideoCrafter}~\cite{chen2023videocrafter}. These models vary in {frame rate}, {resolution}, {duration}, and motion quality, providing a robust sampling of T2V generation characteristics. For instance, \emph{Kling} produces the highest frame rate (30 FPS), while \emph{Dreamina} achieves the highest resolution (1472×832). In contrast, \emph{LVDM} and \emph{Text2Video-Zero} yield the lowest resolution (256×256) and frame rate (4 FPS). Regarding duration, \emph{CogVideo} generates the longest clips (up to 6 seconds), while \emph{Hotshot-XL} and \emph{VideoCrafter} produce shorter sequences (around 1 second). 
In total, we generate {1,130 videos} across 113 prompts, with multiple models applied per prompt.



\subsection{Human Feedback Consolidation}
To enable precise evaluation of the collected AIGVs, each video in the \textit{EduAVQABench} dataset is annotated along two main dimensions:

\noindent \textit{Perceptual Quality.} 
This dimension characterizes video visual fidelity from three complementary perspectives:

\begin{itemize}
    \item \textit{Spatial Quality}: evaluates frame-level fidelity through texture clarity, edge sharpness, and artifact absence.
    \item \textit{Temporal Quality}: assesses motion coherence across frames, considering smoothness and temporal stability.
    \item \textit{Overall Perceptual Quality}: reflects a holistic impression integrating spatial and temporal aspects.
\end{itemize}

\noindent \textit{Prompt Alignment.} 
This dimension measures the semantic consistency between the textual prompt and the generated video content:

\begin{itemize}
    \item \textit{Word-Level Alignment}: examines whether individual keywords or visual entities specified in the prompt are accurately represented in the video.
    \item \textit{Sentence-Level (Overall) Alignment}: evaluates how well the overall visual semantics correspond to the intended meaning of the prompt.
\end{itemize}

 \noindent A total of 19 trained annotators participated in the labeling process. Before annotation, all participants completed a structured training session using a held-out set of videos to calibrate their perception of visual distortions and semantic alignment. The annotation procedure followed the ITU-R BT.500 recommendations for subjective VQA. Each video was independently rated across all five dimensions using a 5-point Likert scale (1: very poor, 5: excellent), ensuring comprehensive and unbiased human evaluation across both perceptual and semantic aspects.

\begin{table}[htp]
\centering
\vspace{-4mm}
\caption{Annotation consistency analysis across evaluators.}
\vspace{-2mm}
\resizebox{\linewidth}{!}{
\renewcommand{\arraystretch}{1.2}
\setlength{\tabcolsep}{3.8pt}
\begin{tabular}{lccc}
\toprule
\textbf{Metric} & \textbf{Avg. SRCC} & \textbf{Avg. PLCC} & \textbf{\#Annotators (SRCC $>$ 0.8)} \\
\midrule
Perceptual Quality & 0.810 & 0.803 & 12 \\
Prompt Alignment & 0.759 & 0.762 & 9 \\
\bottomrule
\end{tabular}}
\label{tab:annotator_consistency}
\end{table}

\label{sec:Annotation Processing and Analysis}
To ensure annotation quality and consistency, we implemented a multi-step post-processing procedure following the ITU-R BT.500 recommendation. After collecting raw ratings from all participants, we perform outlier removal at the individual rating level to ensure annotation quality and consistency. For each video-dimension pair, given a set of scores $\{x_i\}_{i=1}^N$, we compute the sample mean $\mu$ and standard deviation $\sigma$, and define the inlier set as
\begin{equation}
\mathcal{I}_{\text{inlier}} = \left\{ i \,\middle|\, |x_i - \mu| \leq \lambda \cdot \sigma \right\},
\end{equation}
where the threshold $\lambda$ is set to $2.0$ for approximately Gaussian-distributed dimensions, and to $\sqrt{20}$ for dimensions exhibiting significant deviations from normality. Ratings falling outside $\mathcal{I}_{\text{inlier}}$ are excluded from aggregation. Furthermore, annotators with more than $5\%$ of their ratings identified as outliers across all evaluations are removed from subsequent analysis. Annotators are retrained with additional guidance and example-based calibration before returning to the annotation pipeline. 

We further evaluate annotation consistency by measuring the agreement between each annotator and the aggregated MOS, as summarized in Tab.~\ref{tab:annotator_consistency}. The results demonstrate strong agreement across annotators, with average SRCC/PLCC values of 0.810/0.803 for perceptual quality and 0.759/0.762 for prompt alignment. Moreover, 12 annotators achieve SRCC scores above 0.8 in perceptual quality assessment, further validating the reliability of the annotations. More details on dataset construction can be found in the supplementary.

\subsection{Annotation Processing and Analysis}

\begin{figure}[htp]
  \centering
  \includegraphics[width=0.5\textwidth]{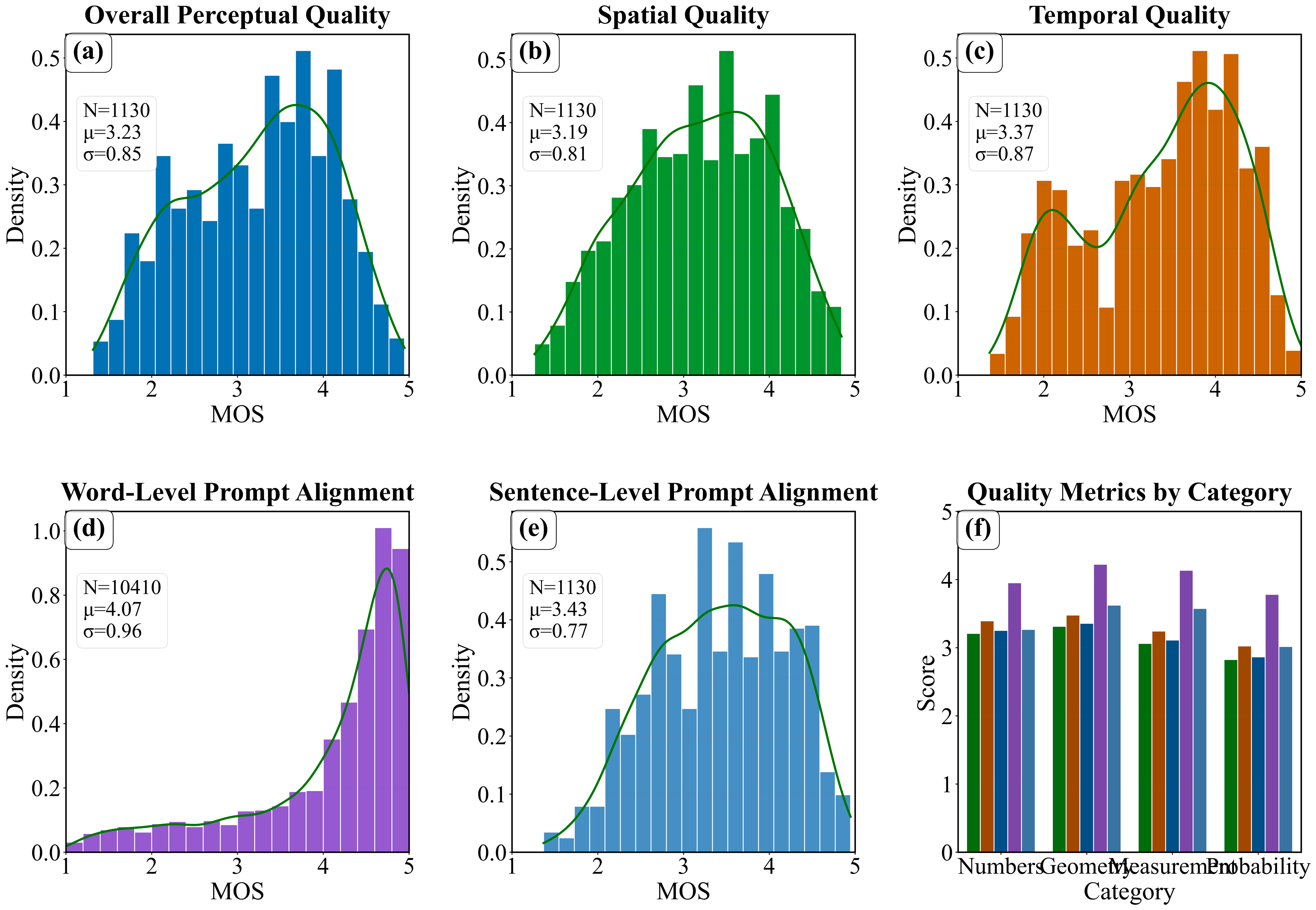}
 \vspace{-5mm}
 \caption{Annotation Analysis. (a)-(e): MOS distributions across five dimensions;
(f): Average MOS of each scene category.}
  \label{fig:MOS}
  \vspace{-5mm}
\end{figure}

The distributions of MOS across the five quality dimensions are shown in Fig.~\ref{fig:MOS}(a)-(e). In particular, the \textit{Spatial Quality}, \textit{Overall Perceptual Quality}, and \textit{Sentence-Level Prompt Alignment} dimensions exhibit approximately Gaussian distributions centered around moderate values, suggesting consistent perceptual fidelity and general semantic correspondence in the majority of videos. By contrast, the \textit{Temporal Quality} shows a distinctly bimodal distribution, reflecting the dual nature of model behavior: either producing temporally stable but static scenes or attempting complex motion that often results in temporal artifacts and lower scores. The \textit{Word-Level Prompt Alignment} dimension is skewed toward higher scores but retains a heavy tail toward lower values. This indicates that while many key terms are visually grounded, certain abstract  remain difficult for current T2V models to represent accurately.  The average MOS  for each scene category is shown in Fig.~\ref{fig:MOS}(f).

\begin{figure*}[t]
    \centering
    \includegraphics[width=1.0\linewidth]{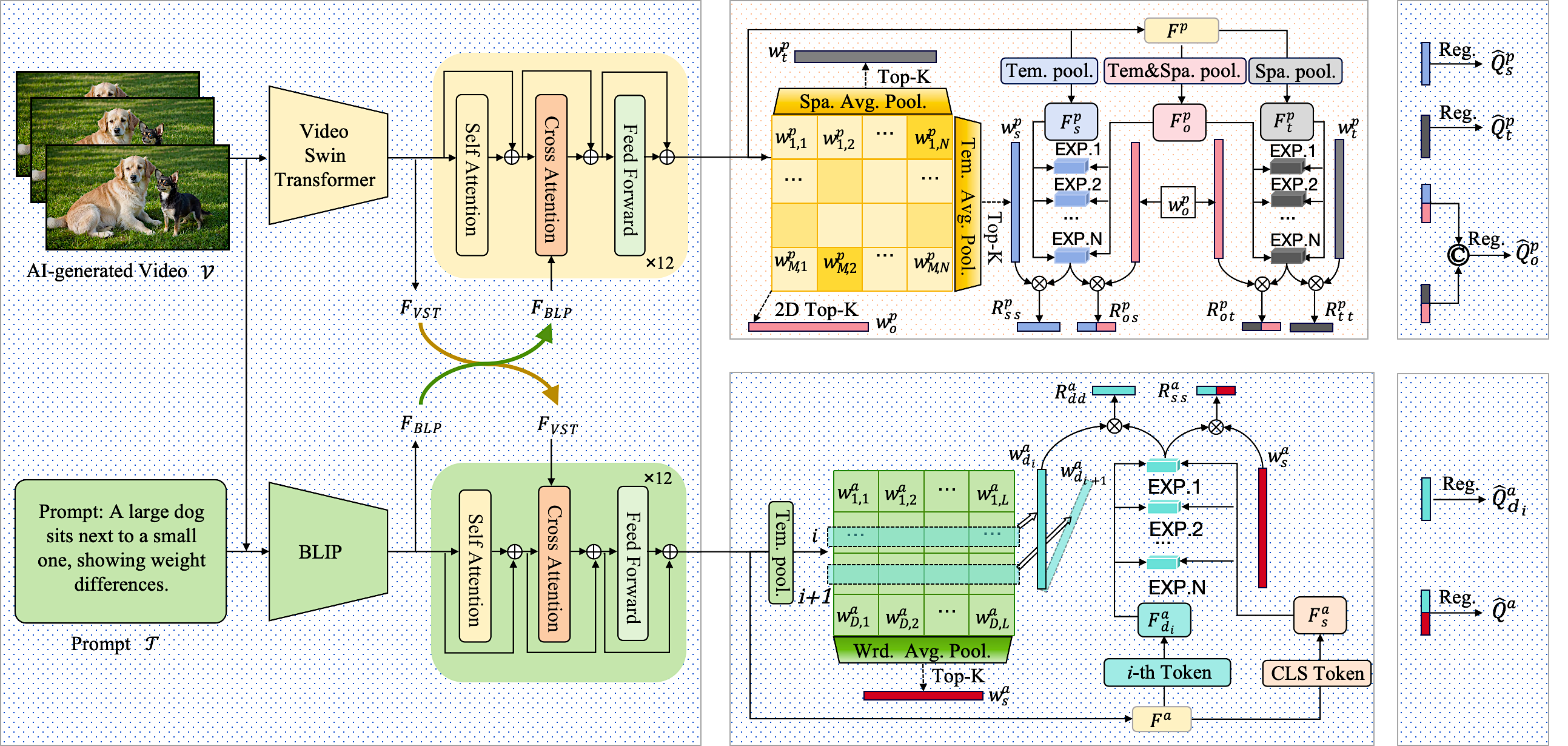}
    \vspace{-4mm}
    \caption{Overview of \textit{EduVQA} framework. We jointly predict five quality dimensions via a dual-path framework equipped with 2D MoE.}
    \label{fig:framework}
    \vspace{-3mm}
\end{figure*}

\section{EduVQA: Multi-dimensional Quality Assessment for AIGV}

\subsection{Overview}

As illustrated in Fig.~\ref{fig:framework}, our proposed {EduVQA} aims to provide fine-grained and interpretable VQA for AIGVs, jointly predicting \emph{perceptual quality} and \emph{prompt alignment quality}. 
EduVQA adopts a {dual-path architecture}, where the perceptual path models low-level distortions (spatial and temporal), while the  alignment path evaluates text–video correspondence at both word and sentence levels.
\paragraph{Feature Extraction.}
Given an input video $\mathcal{V}$ and its corresponding prompt $\mathcal{T}$, our goal is to extract both perceptually sensitive and semantically aligned representations. 
Following T2VQA~\cite{Kou2024GAIA}, we employ a Video Swin Transformer~\cite{liu2022video} to extract video quality-sensitive features, and utilize BLIP~\cite{li2022blip} as a multimodal encoder to obtain fused visual–textual features:
\begin{equation}
F_{VST} = \Phi_{\text{VST}}(\mathcal{V}), \quad F_{VST} \in \mathbb{R}^{T \times H \times W \times C},
\end{equation}
\begin{equation}
F_{BLIP} = \Phi_{\text{BLIP}}(\mathcal{V}, \mathcal{T}), \quad F_{BLIP} \in \mathbb{R}^{T \times L \times C},
\end{equation}
where $T$, $H$, $W$ denote the temporal and spatial dimensions, and $L$ is the token length of the input text.
\paragraph{Cross-modal interaction.}
To aggregate multimodal semantics, two cross-attention operations are designed in opposite directions. 
When generating perceptual-aware features $F_p$, we use $F_{VST}$ as the \textit{query} and $F_{BLIP}$ as the \textit{key}:
\begin{equation}
F_{p} = \text{CrossAttn}(F_{VST}, F_{BLIP}),
\end{equation}
while for alignment features $F_{a}$, the direction is reversed:
\begin{equation}
F_{a} = \text{CrossAttn}(F_{BLIP}, F_{VST}).
\end{equation}

\paragraph{Philosophy of Structured 2D Mixture-of-Experts (S2D-MoE).}
In the subsequent dual-path prediction stage, we incorporate S2D-MoE to capture and strengthen the hierarchical relationships between overall and sub-dimensional quality predictions within each path. 
Compared to conventional MoE layers that activate experts independently for each task, S2D-MoE introduces two key mechanisms: \textit{expert sharing} and an adaptive \textit{2D gating matrix}.

The shared expert pool tightly couples the representation learning of the overall-quality with that of each sub-dimension by requiring them to rely on a common set of experts. This design ensures that the overall-quality representation is intrinsically grounded in features that remain sensitive to each sub-dimension, thereby preventing the overall predictor from diverging from the fine-grained quality semantics.
Building upon this shared foundation, the adaptive 2D gating matrix further models the hierarchical dependence between tasks by integrating the sub-dimension–specific expert weights into the overall gating distribution. As a result, the overall prediction becomes an explicit aggregation of sub-dimension contributions. Collectively, these mechanisms establish a coherent representational structure in which overall quality inference is consistently aligned with all sub-dimensions, ultimately improving interpretability and generalization in multi-dimensional quality assessment.

\subsection{Perceptual Quality Path}

Given perceptual features $F_p \in \mathbb{R}^{T \times H \times W \times C}$, our goal is to estimate spatial, temporal, and overall perceptual quality scores.
We first generate a 2D adaptive gating matrix $\mathbf{W}^p \in \mathbb{R}^{M \times N}$ from the $F_p$, representing the interaction importance between $M$ spatial and $N$ temporal experts.

\noindent \textbf{Structured expert activation.}
We derive three sets of activation weights by pooling $\mathbf{W}^p$:
\begin{align}
W_s^p &= \text{TopK}\big(\text{Mean}_{\text{row}}(\mathbf{W}^p)\big), \\
W_t^p &= \text{TopK}\big(\text{Mean}_{\text{col}}(\mathbf{W}^p)\big), \\
W_o^p &= \text{TopK}\big(\mathbf{W}^p\big).
\end{align}
$W_s^p$ and $W_t^p$ correspond to spatial and temporal gating weight, while $W_o^p$ represents globally activated weights.

\noindent \textbf{Feature aggregation.}
We perform average pooling along temporal and spatial dimensions on $F_p$ to obtain:
\begin{align}
F_s^p &= \text{Pool}_T(F_p) \in \mathbb{R}^{H \times W \times C}, \\
F_t^p &= \text{Pool}_{H,W}(F_p) \in \mathbb{R}^{T \times C}, \\
F_o^p &= \text{Pool}_{T,H,W}(F_p) \in \mathbb{R}^{C}.
\end{align}
Each pooled feature is then processed by the corresponding spatial and temporal expert sets $\{E_j^s\}_1^M$,$\{E_j^t\}_1^N$:
{\small
\begin{align}
R_{ss}^p &= \sum_{j=1}^M W_{s,j}^p \, E_j^s(F_s^p), \quad
R_{tt}^p = \sum_{j=1}^N W_{t,j}^p \, E_j^t(F_t^p), \\
R_{os}^p &= \sum_{j=1}^M W_{o,j}^p \, E_j^s(F_o^p), \quad
R_{ot}^p = \sum_{j=1}^N W_{o,j}^p \, E_j^t(F_o^p).
\end{align}}

\noindent \textbf{Quality regression.}
After feature refinement, we obtain sub-quality and overall quality predictions as:
\begin{align}
\hat{Q}_s^p &= \text{FC}_s(R_{ss}^p), \\
\hat{Q}_t^p &= \text{FC}_t(R_{tt}^p), \\
\hat{Q}_o^p &= \text{FC}_o(R_{os}^p \text{\textcircled{c}} R_{ot}^p),
\end{align}
where $FC$ denotes a fully connected layer, and `$\text{\textcircled{c}}$' represents concatenation along the channel dimension.

\subsection{Prompt Alignment Path}

The alignment branch operates on the multimodal representation $F_a \in \mathbb{R}^{T \times D \times C}$, where $D$ denotes the number of textual tokens.
Following the same philosophy of S2D-MoE, we compute an adaptive gating matrix $\mathbf{W}^a \in \mathbb{R}^{D \times L}$, where $L$ represents the number of semantic experts.

\noindent \textbf{Token-wise and sentence-wise experts.}
For each word token $d_i$, we obtain its expert weights:
\begin{equation}
W_{d_i}^a = \text{TopK}(\mathbf{W}^a[i,:]),
\end{equation}
sentence-level weights are derived by averaging across tokens:
\begin{equation}
W_s^a = \text{TopK}(\text{Mean}_{i}(\mathbf{W}^a[i,:])).
\end{equation}

\noindent \textbf{Feature projection and regression.}
We extract token-level and global textual features:
\begin{equation}
F_{d_i}^a = F_a[:,i,:], \quad F_s^a = F_a[:,\text{CLS},:],
\end{equation}
and process them via the alignment expert set $\{E_j^a\}_1^Z$:
{\small
\begin{align}
R_{dd}^a &= \sum_{j=1}^Z W_{d_i,j}^a E_j^a(F_{d_i}^a), 
R_{ss}^a = \sum_{j=1}^Z W_{s,j}^a E_j^a(F_s^a),
\end{align}
}
leading to the final word-wise ($\hat{Q}_{d_i}^a$) and sentence-wise ($\hat{Q}_a $) alignment predictions:
\begin{equation}
\hat{Q}_{d_i}^a = \text{FC}_d(R_{dd}^a), \quad
\hat{Q}^a = \text{FC}_s(R_{ss}^a).
\end{equation}

\subsection{Multi-task Optimization}

The entire EduVQA is optimized under a  multi-task objective $\mathcal{L}_{\text{total}}$:
\begin{equation}
\begin{aligned}
\mathcal{L}_{\text{total}} =&~ \lambda_1 \mathcal{L}_{plcc}(\hat{Q}_s^p, Q_s^p)
+ \lambda_2 \mathcal{L}_{plcc}(\hat{Q}_t^p, Q_t^p)\\ &+ \lambda_3 \mathcal{L}_{plcc}(\hat{Q}_o^p, Q_o^p) + \lambda_4 \frac{1}{L} \sum_{i=1}^{L}  \mathcal{L}_{plcc}(\hat{Q}_{d_i}^a, Q_{d_i}^a)\\
&+ \lambda_5 \mathcal{L}_{plcc}(\hat{Q}^a, Q^a),
\label{all}
\end{aligned}
\end{equation}
where $\mathcal{L}_{plcc}$ denotes the Pearson Linear Correlation Coefficient loss~\cite{wu2022fast}.
$Q_s^p$, $Q_t^p$, and $Q_o^p$ are the ground-truth spatial, temporal, and overall perceptual quality scores.
$Q_{d_i}^a$ and $Q^a$ represent the ground-truth $i$-th word-level and sentence-level alignment quality, respectively.
$L$ is the number of word tokens, and $\lambda_1$–$\lambda_5$ balance the contributions of the sub-tasks.
This joint optimization promotes interaction between sub-dimensional and overall predictions, enabling EduVQA to deliver accurate and interpretable prediction.

\section{Experiments}
\subsection{Implementation details} 
We conduct experiments on the EduAVQABench dataset, split into training, validation, and test sets with a ratio of 6:2:2. To ensure fair comparison, we perform 10 random splits based on video categories and generative models. For each split, we ensure that the number of videos generated by each model and from each category is balanced. Final performance is reported as the average across these 10 splits. Each expert is implemented as a two-layer multilayer perceptron (MLP) with ReLU activation. For each sub-dimension, we employ 8 experts ($M$=$N$=$Z$=8) and all dimension select  top-2 experts during routing. 
The model is optimized by the Adam optimizer with an initial learning rate of 1e-5. The $\lambda_1$ to $\lambda_5$ in Eqn.~(\ref{all}) are set by 0.125, 0.125, 0.25, 0.25, and 0.25, respectively. A cosine annealing schedule is used to gradually reduce the learning rate to zero over the course of training. We train the model for 50 epochs with a batch size of 4.

\begin{table}[htp]
\centering
\vspace{-3mm}
\caption{Comparison on \textit{Perceptual Quality} and \textit{Prompt Alignment} of EduAVQABench. \textcolor{red}{$\uparrow$}: higher is better, \textcolor{purple}{$\downarrow$}: lower is better.}
\vspace{-2mm}
\label{tab:main_results_single}
\resizebox{\linewidth}{!}{
\begin{tabular}{lllcccc}
\toprule
\textbf{Metric} & \textbf{Setting} & \textbf{Method} & \textbf{SRCC}~\textcolor{red}{$\uparrow$} & \textbf{PLCC}~\textcolor{red}{$\uparrow$} & \textbf{KRCC}~\textcolor{red}{$\uparrow$} & \textbf{RMSE}~\textcolor{purple}{$\downarrow$}\\
\midrule

\multirow{13}{*}{\rotatebox{90}{\textit{Perceptual Quality}}} 

 & \multirow{2}{*}{\textit{Zero-shot}} 
   & ImageReward~\cite{xu2023imagereward} & 0.409 & 0.432 & 0.282 & 3.369 \\
 &  & Q-Align~\cite{wu2023q} & 0.644 & 0.669 & -- & -- \\
 \cmidrule(lr){2-7}
 & \multirow{10}{*}{\textit{Fine-tuned}} 
   & IPCE~\cite{peng2024aigc} & 0.822 & 0.822 & 0.631 & 0.494 \\
 &  & IP-IQA~\cite{qu2024bringing} & 0.852 & 0.863 & 0.666 & 0.434 \\
 &  & DOVER~\cite{Wu2023BVQA} & 0.832 & 0.840 & 0.645 & 0.479 \\
 &  & BVQA~\cite{Li2022Blindly} & 0.848 & 0.862 & -- & -- \\
 &  & CLIPVQA~\cite{xing2024clipvqa} & 0.846 & 0.832 & -- & -- \\
 &  & FasterVQA~\cite{wu2022fast} & 0.844 & 0.856 & 0.659 & 0.456 \\
 &  & GSTVQA~\cite{chen2021learning} & 0.837 & 0.849 & 0.655 & 0.456 \\
 &  & VSFA~\cite{li2019quality} & 0.803 & 0.805 & 0.611 & 0.515 \\
 &  & SimpleVQA~\cite{cheng2025simplevqa} & 0.742 & 0.758 & 0.552 & 0.553 \\
 &  & T2VQA~\cite{Kou2024GAIA} & 0.810 & 0.826 & 0.623 & 0.499 \\
 \cmidrule(lr){2-7}
 \rowcolor{gray!15}
 & \textbf{Ours} & \textbf{EduVQA} & \textbf{0.869} & \textbf{0.879} & \textbf{0.688} & \textbf{0.417} \\

\midrule

\multirow{8}{*}{\rotatebox{90}{\textit{Prompt Alignment}}} 

 & \multirow{4}{*}{\textit{Zero-shot}} 
   & CLIP~\cite{wang2023exploring} & 0.270 & 0.282 & 0.184 & 3.174 \\
 &  & ImageReward~\cite{xu2023imagereward} & 0.409 & 0.432 & 0.282 & 3.369 \\
 &  & BLIP~\cite{li2022blip} & 0.457 & 0.392 & 0.317 & 101.113 \\
 &  & Open-VCLIP~\cite{weng2023open} & 0.335 & 0.229 & 0.340 & 94.306 \\
 \cmidrule(lr){2-7}
 & \multirow{3}{*}{\textit{Fine-tuned}} 
   & IPCE~\cite{peng2024aigc} & 0.634 & 0.643 & 0.460 & 0.604 \\
 &  & IP-IQA~\cite{qu2024bringing} & 0.648 & 0.653 & 0.469 & 0.593 \\
 &  & T2VQA~\cite{Kou2024GAIA} & 0.731 & 0.733 & 0.544 & 0.564 \\
 \cmidrule(lr){2-7}
 \rowcolor{gray!15}
 & \textbf{Ours} & \textbf{EduVQA} & \textbf{0.757} & \textbf{0.768} & \textbf{0.571} & \textbf{0.527} \\

\bottomrule
\end{tabular}
}
\label{tab:edu1k}
\end{table}
\vspace{-7mm}

\subsection{Baselines}
To comprehensively benchmark our approach, we compare against a wide range of baselines across the two key quality dimensions: \textit{(1) Perceptual Quality Baselines.}
We evaluate several image and video quality assessment models. ImageReward~\cite{xu2023imagereward}, IPCE~\cite{peng2024aigc}, and IP-IQA~\cite{qu2024bringing} are originally developed for image-level AIGC quality assessment. 
For video-oriented baselines, we consider DOVER~\cite{Wu2023BVQA}, BVQA~\cite{Li2022Blindly}, CLIPVQA~\cite{xing2024clipvqa}, FasterVQA~\cite{wu2022fast}, GSTVQA~\cite{chen2021learning}, VSFA~\cite{li2019quality}, and SimpleVQA~\cite{cheng2025simplevqa}, which are primarily designed for user-generated content (UGC) quality assessment. We fine-tune these models on our EduAVQABench dataset for fair comparison. Q-Align is evaluated in a zero-shot setting using publicly released weights. In addition, we include T2VQA, a recent method specifically designed for evaluating T2V generation models. \textit{(2) Prompt Alignment Baselines.}
We evaluate several vision-language models with strong zero-shot alignment capabilities, including CLIP~\cite{wang2023exploring}, ImageReward~\cite{xu2023imagereward}, BLIP~\cite{li2022blip}, and Open-VCLIP~\cite{weng2023open}, using their pre-trained weights without fine-tuning. 
For image-based models, we aggregate the frame-wise results via average pooling.

\subsection{Quantitative Analysis}
In Tab.~\ref{tab:edu1k}, we present a comprehensive comparison across all baseline methods on the EduAVQABench dataset. Our proposed model, \textit{EduVQA}, consistently outperforms existing methods in both perceptual quality and prompt alignment evaluation, demonstrating strong generalization capability for educational AIGC content. For perceptual quality evaluation, \textit{EduVQA} achieves the best performance across all metrics, outperforming the state-of-the-art image-level model \textit{IP-IQA} by +2.00\% (SRCC), and +1.85\% (PLCC). Notably, unlike image-based models such as \textit{IP-IQA}, which process each frame independently, failing to capture temporal dynamics, our video-level framework inherently models temporal dependencies. Compared to the strongest video-level baseline \textit{BVQA}, \textit{EduVQA} also shows consistent improvements of +2.48\% in SRCC.  In terms of prompt alignment, \textit{EduVQA} significantly outperforms the most competitive fine-tuned model, \textit{T2VQA}, achieving a +3.56\% SRCC, and +4.77\% PLCC improvement. These results highlight the superior capability of our \textit{EduVQA} in modeling fine-grained video-text alignment beyond static frame-level correlations. In contrast, zero-shot models such as CLIP, BLIP, and ImageReward perform worse, further emphasizing the importance of domain adaptation for this task.

\begin{table}[htp]
\centering
\vspace{-4mm}
\caption{Cross-dataset performance comparison. All models are trained on our EduAVQABench dataset.}
\vspace{-2mm}
\resizebox{\linewidth}{!}{
\renewcommand{\arraystretch}{1.2}
\setlength{\tabcolsep}{3.8pt}
\begin{tabular}{l|ccc|cc}
\toprule

\multirow{3}{*}{\textbf{Method}} 
& \multicolumn{3}{c|}{\textbf{LGVQ~\cite{Zhang2024BenchmarkingMA} Dataset}} 
& \multicolumn{2}{c}{\textbf{EvalCrafter~\cite{liu2024evalcrafter} Dataset}} \\ 

\cmidrule(lr){2-6}
& \textbf{Spatial} & \textbf{Temporal} & \textbf{ Alignment} 
& \textbf{Perceptual} & \textbf{Alignment} \\ 

\cmidrule(lr){2-6}
& \textbf{SRCC / PLCC} & \textbf{SRCC / PLCC} & \textbf{SRCC / PLCC} & \textbf{SRCC / PLCC} & \textbf{SRCC / PLCC} \\ 

\midrule
BVQA~\cite{Li2022Blindly}           & 0.518 / \textbf{0.608} & 0.235 / 0.550 & -- / -- & 0.161 / 0.172 & -- / -- \\
CLIPVQA~\cite{xing2024clipvqa}      & 0.509 / 0.480 & 0.355 / 0.287 & -- / -- & 0.319 / 0.325 & -- / -- \\
FasterVQA~\cite{wu2022fast}& 0.511 / 0.539 & 0.380 / 0.427 & -- / -- & 0.378 / 0.359 & -- / -- \\
GSTVQA~\cite{chen2021learning}      & 0.507 / 0.555 & 0.459 / 0.529 & -- / -- & 0.276 / 0.281 & -- / -- \\
VSFA~\cite{li2019quality}           & 0.498 / 0.516 & 0.174 / 0.246 & -- / -- & 0.213 / 0.195 & -- / -- \\
SimpleVQA~\cite{cheng2025simplevqa} & 0.395 / 0.442 & 0.262 / 0.394 & -- / -- & 0.268 / 0.289 & -- / -- \\
IPCE~\cite{peng2024aigc}            & 0.324 / 0.336 & 0.428 / 0.473 & 0.201 / 0.190 & 0.247 / 0.255 & 0.245 / 0.253 \\
IP-IQA~\cite{qu2024bringing}        & 0.423 / 0.473 & 0.363 / 0.454 & 0.267 / 0.256 & 0.242 / 0.239 & 0.131 / 0.137 \\
T2VQA~\cite{Kou2024GAIA}            & 0.466 / 0.504 & 0.402 / 0.453 & 0.521 / 0.539 & 0.305 / 0.299 & 0.553 / 0.554 \\
\midrule
\rowcolor{gray!15}
\textbf{EduVQA (Ours)}              & \textbf{0.536} / 0.593 & \textbf{0.511} / \textbf{0.571} & \textbf{0.529} / \textbf{0.547} & \textbf{0.408} / \textbf{0.398} & \textbf{0.586} / \textbf{0.599} \\
\bottomrule
\end{tabular}}
\label{tab:cross}
\end{table}

\vspace{-2mm}

\begin{table}[htp]
\centering
\vspace{-4mm}
\caption{Cross-dataset training analysis using different training datasets. 
All models are evaluated on the same target benchmark without fine-tuning.}
\vspace{-2mm}
\resizebox{\linewidth}{!}{
\renewcommand{\arraystretch}{1.2}
\setlength{\tabcolsep}{3.8pt}
\begin{tabular}{l|l|cc|cc}
\toprule

\multirow{2}{*}{\textbf{Train Dataset}} & 
\multirow{2}{*}{\textbf{Test Dataset}} & 
\multicolumn{2}{c|}{\textbf{Perceptual Quality}} & 
\multicolumn{2}{c}{\textbf{Prompt Alignment}} \\

\cmidrule(lr){3-6}

& & \textbf{SRCC} & \textbf{PLCC} & \textbf{SRCC} & \textbf{PLCC} \\
\midrule  

\rowcolor{gray!15}
\textbf{EduAVQABench (Ours)} & LGVQ~\cite{Zhang2024BenchmarkingMA} & \textbf{0.5558} & \textbf{0.6346} & \textbf{0.5745} & \textbf{0.5941} \\
AIGVE-Bench~\cite{xiang2025aigvetoolaigeneratedvideoevaluation} & LGVQ~\cite{Zhang2024BenchmarkingMA} & 0.2111 & 0.1986 & 0.0759 & 0.0777 \\
TVGE~\cite{wu2024bettermetrictexttovideogeneration} & LGVQ~\cite{Zhang2024BenchmarkingMA} & 0.1488 & 0.1688 & 0.4950 & 0.5223 \\
\midrule  

\rowcolor{gray!15}
\textbf{EduAVQABench (Ours)} & TVGE~\cite{wu2024bettermetrictexttovideogeneration} & \textbf{0.3950} & \textbf{0.3866} & 0.5580 & 0.5343 \\
AIGVE-Bench~\cite{xiang2025aigvetoolaigeneratedvideoevaluation} & TVGE~\cite{wu2024bettermetrictexttovideogeneration} & 0.2532 & 0.2630 & 0.1772 & 0.1685 \\
LGVQ~\cite{Zhang2024BenchmarkingMA} & TVGE~\cite{wu2024bettermetrictexttovideogeneration} & 0.3796 & 0.3634 & \textbf{0.5659} & \textbf{0.5373} \\
\midrule  

\rowcolor{gray!15}
\textbf{EduAVQABench (Ours)} & Q-Eval-100K~\cite{zhang2025qeval100kevaluatingvisualquality} & \textbf{0.3374} & \textbf{0.3405} & 0.4358 & 0.4456 \\
AIGVE-Bench~\cite{xiang2025aigvetoolaigeneratedvideoevaluation} & Q-Eval-100K~\cite{zhang2025qeval100kevaluatingvisualquality} & 0.1609 & 0.1566 & 0.0914 & 0.0980 \\
TVGE~\cite{wu2024bettermetrictexttovideogeneration} & Q-Eval-100K~\cite{zhang2025qeval100kevaluatingvisualquality} & 0.2080 & 0.1948 & \textbf{0.4522} & 0.4486 \\
LGVQ~\cite{Zhang2024BenchmarkingMA} & Q-Eval-100K~\cite{zhang2025qeval100kevaluatingvisualquality} & 0.2889 & 0.2965 & 0.4320 & \textbf{0.4502} \\
\midrule  

\rowcolor{gray!15}
\textbf{EduAVQABench (Ours)} & AIGVE-60K~\cite{wang2025lovebenchmarkingevaluatingtexttovideo} & \textbf{0.5902} & 0.5704 & 0.5428 & 0.5627 \\
AIGVE-Bench~\cite{xiang2025aigvetoolaigeneratedvideoevaluation} & AIGVE-60K~\cite{wang2025lovebenchmarkingevaluatingtexttovideo} & 0.1926 & 0.1864 & 0.1796 & 0.1832 \\
TVGE~\cite{wu2024bettermetrictexttovideogeneration} & AIGVE-60K~\cite{wang2025lovebenchmarkingevaluatingtexttovideo} & 0.3063 & 0.3147 & \textbf{0.5683} & 0.5592 \\
LGVQ~\cite{Zhang2024BenchmarkingMA} & AIGVE-60K~\cite{wang2025lovebenchmarkingevaluatingtexttovideo} & 0.5677 & \textbf{0.5742} & 0.5394 & \textbf{0.5694} \\
\bottomrule

\end{tabular}}
\label{tab:cross_dataset_training}
\end{table}
\vspace{-5mm}


\subsection{Cross-dataset Evaluation}
To evaluate the generalization capability of our \textit{EduVQA}, we conduct cross-dataset testing as shown in Tab.~\ref{tab:cross}. All methods are trained on EduAVQABench dataset and directly tested on two unseen AIGVQA benchmarks, LGVQ~\cite{Zhang2024BenchmarkingMA} and EvalCrafter~\cite{liu2024evalcrafter}, without fine-tuning. Our model consistently surpasses existing approaches. On LGVQ, it achieves the highest SRCC in both spatial and temporal dimensions (0.536 and 0.511), demonstrating strong robustness to domain shift. It also establishes new state-of-the-art results in prompt alignment, confirming its superior understanding of text-conditioned semantics. When transferred to EvalCrafter, our model maintains leading performance in both quality dimensions.
These results validate both the representational strength of EduAVQABench and the robustness of our model design.

To further evaluate the effectiveness of \textit{EduAVQABench}, we additionally conduct cross-dataset training analysis as shown in Tab.~\ref{tab:cross_dataset_training}. Specifically, we train \textit{EduVQA} on different AIGVQA datasets, including AIGVE-Bench~\cite{xiang2025aigvetoolaigeneratedvideoevaluation}, TVGE~\cite{wu2024bettermetrictexttovideogeneration}, LGVQ~\cite{Zhang2024BenchmarkingMA}, and our EduAVQABench, and evaluate them on the same target benchmarks without fine-tuning. 
Compared with existing datasets, EduAVQABench contains 1,130 videos with 310,650 fine-grained annotations, resulting in a significantly higher annotation density despite its smaller video scale (\eg ``AIGVE-Bench: 2,430 videos / 21,870 annotations; TVGE: 2,543 videos / 50,860 annotations; LGVQ: 2,808 videos / 84,240 annotations''). This design emphasizes dense multi-dimensional supervision rather than sheer video quantity.
As shown in the table, models trained on EduAVQABench consistently outperform those trained on other datasets across all evaluation settings and test benchmarks. For example, when tested on LGVQ, EduAVQABench achieves the highest perceptual SRCC/PLCC of 0.5558/0.6346 and alignment SRCC/PLCC of 0.5745/0.5941, significantly surpassing AIGVE-Bench and TVGE. Similar trends are observed on TVGE~\cite{wu2024bettermetrictexttovideogeneration}, Q-Eval-100K~\cite{zhang2025qeval100kevaluatingvisualquality}, and AIGVE-60K~\cite{wang2025lovebenchmarkingevaluatingtexttovideo}, where EduAVQABench shows consistently superior or highly competitive performance across both perceptual quality and prompt alignment metrics. These results demonstrate that the large-scale fine-grained annotations in EduAVQABench provide more effective supervision signals, leading to stronger cross-dataset generalization.

\begin{figure*}[t]
\centering
\includegraphics[width=\linewidth]{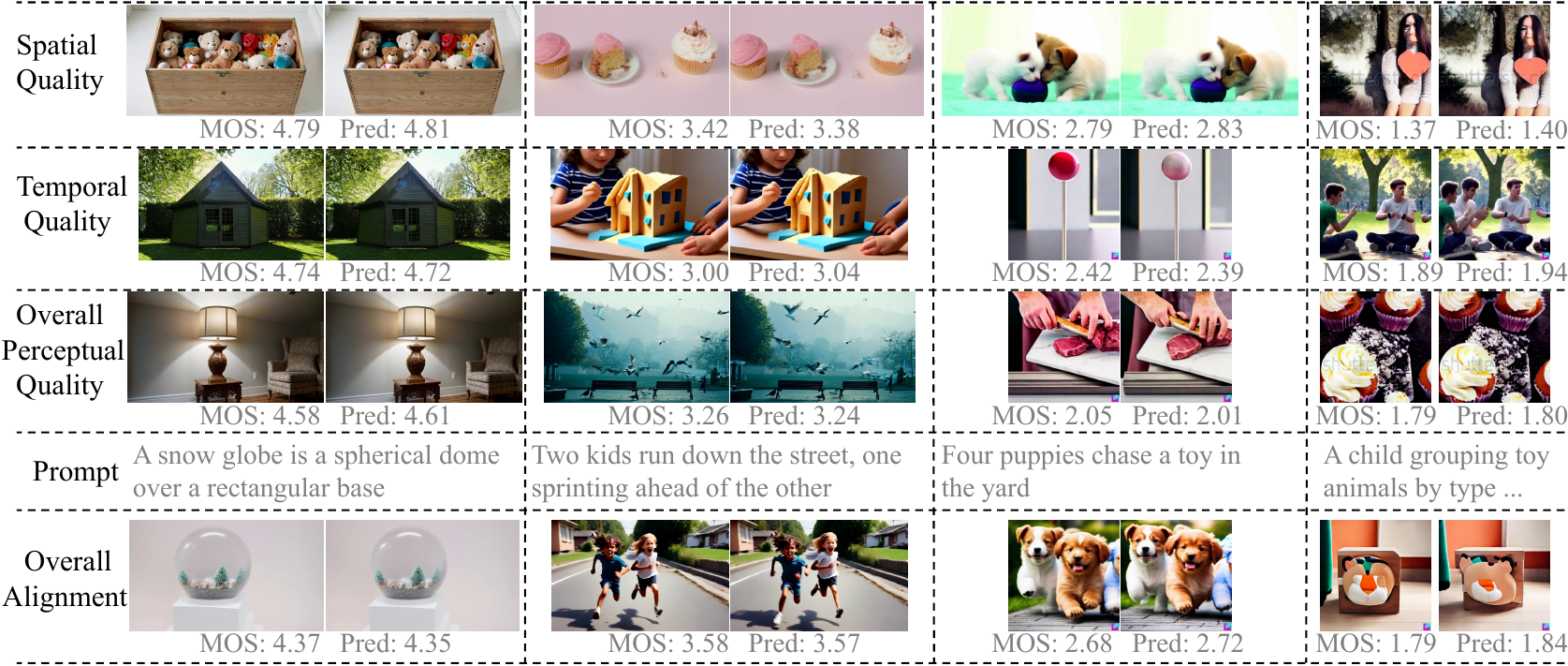}

\vspace{-3mm} 
\caption{
Comparison between predicted scores and MOS across multiple quality dimensions and quality levels.
}
\vspace{-5mm}
\label{fig:dataset_mos_pred}
\end{figure*}


\begin{figure*}[htp]
\centering
\includegraphics[width=\linewidth]{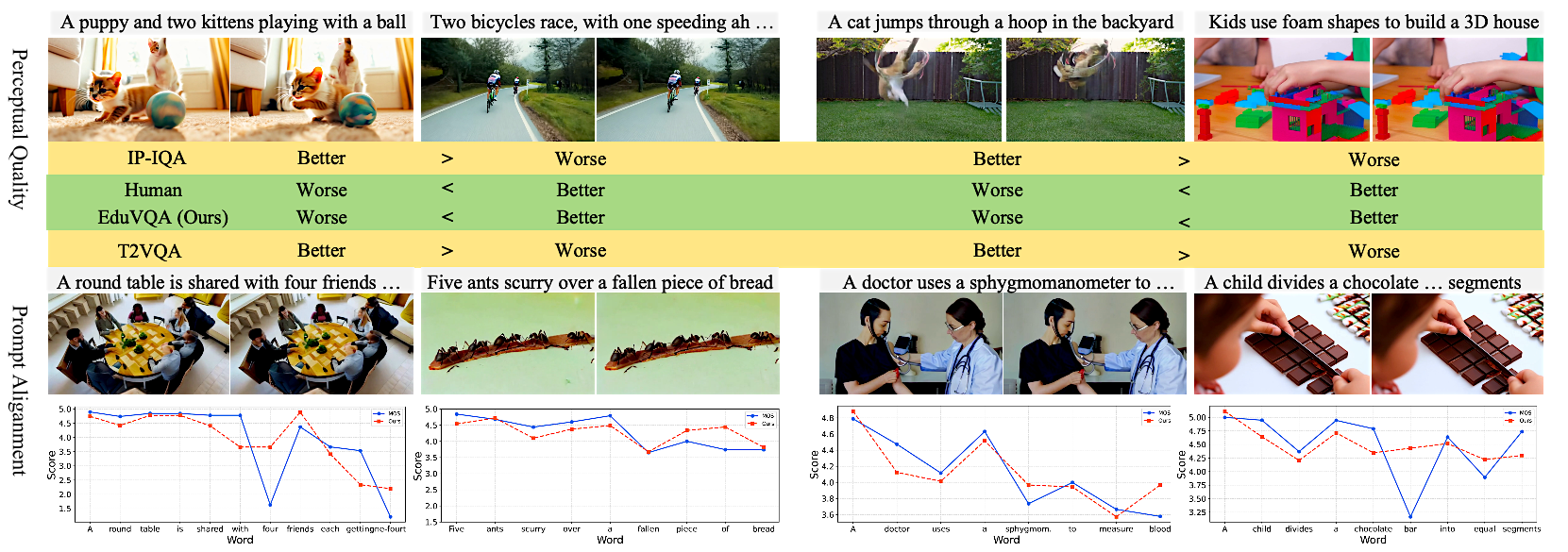}
\vspace{-7mm}
\caption{Qualitative comparison of perceptual quality (top row) and prompt alignment (bottom row). We compare our EduVQA model against state-of-the-art baselines, IP-IQA and T2VQA, in each quality dimensions. In each video pair, the right video exhibits \textit{superior} perceptual quality or prompt alignment compared to the left. EduVQA consistently aligns with human judgments, while IP-IQA and T2VQA produce rankings contrary to the MOS.}
\vspace{-5mm}
\label{fig:qualitative}
\end{figure*}

\begin{figure}[htp]
  \centering
  \includegraphics[width=0.5\textwidth]{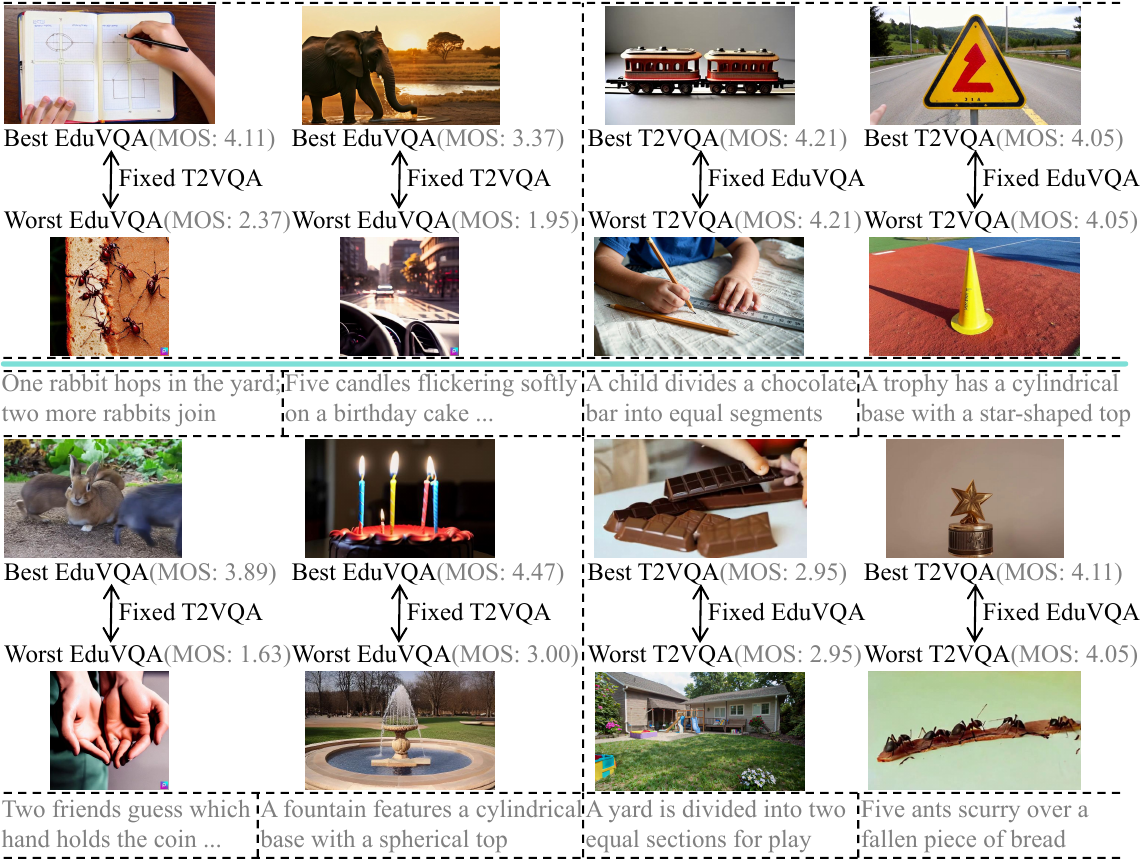}
\vspace{-5mm}
\caption{gMAD competition results between T2VQA and our EduVQA. Top: perceptual quality comparison; Bottom: prompt alignment comparison. zoomed-in view for clarity.}
\vspace{-3mm}
  \label{fig:gmad}
\end{figure}

\subsection{Qualitative Examples} 

In Fig.~\ref{fig:dataset_mos_pred}, we present representative samples across multiple quality dimensions to further examine the relationship between model predictions and human judgments. The results demonstrate that our method reliably tracks perceptual degradations and semantic inaccuracies, producing scores that closely align with human assessment across both spatial–temporal fidelity and prompt grounding. Even in challenging scenarios, such as rapid motion, fine-grained scene details, or abstract mathematical instructions, the predicted quality remains stable, interpretable, and consistent with observed subjective trends.

To further demonstrate the effectiveness of \textit{EduVQA}, we present representative qualitative examples in Fig.~\ref{fig:qualitative}. In the top row, the left videos in each pair exhibit noticeable flickering artifacts and inconsistent motion across frames. While our model accurately reflects the degraded perceptual quality caused by these temporal distortions, IP-IQA, the strongest perceptual-quality baseline, fails to capture such inter-frame inconsistencies, confirming our model captures quality beyond spatial appearance. In the bottom example, although the left videos in each pair appear roughly aligned with the prompt, there is a semantic mismatch with the keywords ``four" or ``measure". Our model precisely identifies the word with weak correspondence, offering fine-grained interpretability. In contrast, T2VQA, the most competitive baseline, tends to produce contradictory rankings, lacking fine-grained assessment capability. More qualitative results are provided in the supplementary material.

Moreover, we perform a gMAD~\cite{ma2018group} competition to compare \textit{EduVQA} and \textit{T2VQA} on both perceptual quality and prompt alignment. 
The gMAD identifies video pairs where one model assigns similar scores while the other produces markedly different predictions, revealing inconsistencies with human perception.
As shown in Fig.~\ref{fig:gmad}, we organize the comparisons into two groups, corresponding to perceptual quality (top) and prompt alignment (bottom). For the perceptual quality group: in columns 1-2, \textit{T2VQA} acts as the defender, assigning similar scores to clips that humans perceive as different; in columns 3-4, \textit{T2VQA} acts as the attacker, predicting different scores for clips that humans perceive as similar. In both cases, \textit{EduVQA} correctly distinguishes or aligns with human perception, demonstrating its sensitivity to subtle temporal distortions. 
Similarly, for the prompt alignment group, we test the defender and attacker setups: \textit{EduVQA} consistently provides scores that better match human judgments, accurately capturing semantic mismatches that \textit{T2VQA} fails to detect. These results highlight the superior robustness of \textit{EduVQA}.

\subsection{Ablation Study}
The contribution of each component is analysed through  ablation studies summarized in Tab.~\ref{tab:ablation}. 
Starting from a baseline without cross-modality fusion or sub-dimensional supervision (ID 1), 
introducing the fusion module (ID 2) yields clear gains, confirming the necessity of visual–textual interaction. 
Adding the spatial–temporal (ST) and word-level (WL) branches (ID 3) further improves both perceptual and alignment correlations. 
Removing either ST (ID 4) or WL (ID 5) leads to performance drops in the corresponding dimension, indicating their complementarity. 
Replacing our specialized MoE with a conventional 1D MoE  (ID 6) also degrades performance, 
demonstrating the advantage of structured 2D expert routing. 
The full configuration (ID 7) achieves the best overall correlation, validating the effectiveness of jointly modeling fusion, expert learning, and fine-grained supervision.

We analyze the impact of the number of experts and Top-K selection in the MoE module, as shown in Tab.~\ref{tab:moe_analysis}.
We first fix Top-K=2 and vary the number of experts. Increasing the number of experts from 4 to 8 leads to clear improvements in both perceptual quality and prompt alignment (from 0.845/0.858 to 0.869/0.879 and from 0.751/0.759 to 0.757/0.768, respectively). However, further increasing the number of experts to 16 does not yield additional gains and instead results in slight performance degradation, suggesting that an overly large expert pool introduces redundancy without improving representation capacity.
We then fix the number of experts to 8 and vary Top-K. We observe that Top-K=2 consistently achieves the best performance, outperforming both sparser routing (Top-K=1) and denser routing (\eg Top-K=4 or 8). This indicates that overly sparse routing limits the model capacity, while activating too many experts weakens specialization and reduces the effectiveness of expert selection.
Overall, these results show that a moderate expert pool combined with sparse routing (8 experts with Top-K=2) provides the best balance between representation capacity and expert specialization, while maintaining computational efficiency.

\begin{table}[htp]
\centering
\vspace{-4mm}
\caption{Ablation results on cross-modality fusion, MoE configurations, and fine-grained sub-dimension supervision. ``ST" and ``WL" denote spatial–temporal perceptual and word-level alignment branches, respectively.}
\vspace{-2mm}
\resizebox{\linewidth}{!}{
\renewcommand{\arraystretch}{1.2}
\setlength{\tabcolsep}{3.8pt}
\begin{tabular}{c|c|ccc|cc|cc|cc}
\toprule
\multirow{2}{*}{\textbf{ID}} & \multirow{2}{*}{\textbf{Fusion}} & 
\multicolumn{3}{c|}{\textbf{MoE}} & 
\multicolumn{2}{c|}{\textbf{Sub-dim.}} & 
\multicolumn{2}{c|}{\textbf{Perception}} & 
\multicolumn{2}{c}{\textbf{Alignment}} \\ 
\cmidrule(lr){3-11}
& & \textbf{Vanilla} & \textbf{Per.} & \textbf{Aln.} & \textbf{ST} & \textbf{WL} & \textbf{PLCC} & \textbf{SRCC} & \textbf{PLCC} & \textbf{SRCC} \\ 
\midrule 
1 & \xmark & \xmark & \xmark & \xmark & \xmark & \xmark & 0.845 & 0.836 & 0.744 & 0.737 \\
2 & \cmark & \xmark & \xmark & \xmark & \xmark & \xmark & 0.863 & 0.851 & 0.756 & 0.750 \\
3 & \cmark & \xmark & \xmark & \xmark & \cmark & \cmark & 0.869 & 0.859 & 0.760 & 0.751 \\
4 & \cmark & \xmark & \xmark & \cmark & \xmark & \cmark & 0.860 & 0.849 & 0.762 & 0.754 \\
5 & \cmark & \xmark & \cmark & \xmark & \cmark & \xmark & 0.873 & 0.862 & 0.747 & 0.741 \\
6 & \cmark & \cmark & \cmark & \cmark & \cmark & \cmark & 0.874 & 0.862 & 0.763 & 0.754 \\
\midrule 
7 & \cmark & \xmark & \cmark & \cmark & \cmark & \cmark & \textbf{0.879} & \textbf{0.869} & \textbf{0.768} & \textbf{0.757} \\
\bottomrule
\end{tabular}
}
\label{tab:ablation}
\end{table}

\vspace{-2mm}

\begin{table}[htp]
\centering
\vspace{-4mm}
\caption{Impact of Expert Number and Top-K Selection in MoE.}
\vspace{-2mm}
\resizebox{\linewidth}{!}{
\begin{tabular}{c|c|cc|cc}
\toprule
\multirow{2}{*}{\textbf{Num. Experts}} & \multirow{2}{*}{\textbf{Top-K}} & 
\multicolumn{2}{c|}{\textbf{Perceptual Quality}} & 
\multicolumn{2}{c}{\textbf{Prompt Alignment}} \\
\cmidrule(lr){3-6}
& & \textbf{SRCC} & \textbf{PLCC} & \textbf{SRCC} & \textbf{PLCC} \\
\midrule
4  & 2 & 0.845 & 0.858 & 0.751 & 0.759 \\
\midrule
8  & 1 & 0.856 & 0.869 & 0.746 & 0.753 \\
\rowcolor{gray!15}
8  & 2 & \textbf{0.869} & \textbf{0.879} & \textbf{0.757} & \textbf{0.768} \\
8  & 4 & 0.850 & 0.863 & 0.737 & 0.745 \\
8  & 8 & 0.853 & 0.864 & 0.756 & 0.761 \\
\midrule
16 & 2 & 0.864 & 0.875 & 0.755 & 0.764 \\
16 & 4 & 0.861 & 0.873 & \textbf{0.757} & 0.764 \\
16 & 8 & 0.852 & 0.865 & 0.744 & 0.755 \\
\bottomrule
\end{tabular}
}
\label{tab:moe_analysis}
\end{table}

\section{Conclusion}
We introduce \textit{EduAVQABench}, the first benchmark dedicated to assessing the quality of AI-generated educational videos.  With fine-grained perceptual and alignment annotations, it enables more reliable and interpretable evaluation of instructional content. 
Built upon this dataset, our \textit{EduVQA} model employs a structured 2D Mixture-of-Experts (MoE) design to jointly capture spatial–temporal fidelity and word-level alignment. 
Comprehensive experiments verify its superior capability in detecting critical quality issues in educational AIGVs. We believe this work establishes a solid foundation for future research on quality-aware generation and evaluation in education-oriented AI video systems.



\bibliographystyle{IEEEtran}
\bibliography{main}

\end{document}